\documentclass[11pt,a4paper]{article}
\usepackage[hyperref]{acl2017}
\usepackage{times}
\usepackage{latexsym}
\usepackage{amsopn}
\usepackage{amssymb}
\usepackage{amsmath}
\usepackage{caption}
\usepackage{subcaption}
\usepackage{url}
\usepackage{graphicx}
\usepackage{hyperref}

\aclfinalcopy 

\newcommand{\ignore}[1]{}

\DeclareMathOperator*{\argmax}{argmax}
\DeclareMathOperator*{\softmax}{softmax}

\title{Emergent Predication Structure\\in Hidden State Vectors of Neural Readers}

\author{Hai Wang\thanks{$\;$ Authors contributed equally.}\quad Takeshi Onishi$^{\ast}$ \quad  Kevin Gimpel\quad David McAllester\\
	Toyota Technological Institute at Chicago \\
	6045 S. Kenwood Ave., Chicago, Illinois 60637, USA \\
	\texttt{\{haiwang,tonishi,kgimpel,mcallester\}@ttic.edu} \\
}

\date{}

\begin{document}
\maketitle
\begin{abstract}

  A significant number of neural architectures for reading comprehension have recently been developed and evaluated on large cloze-style datasets.
  We present experiments supporting the emergence of  ``predication structure'' in the hidden state vectors of these readers.  More specifically, we provide evidence that
  the hidden state vectors represent atomic formulas $\Phi[c]$ where $\Phi$ is a semantic property (predicate) and $c$ is a constant symbol entity identifier.

\end{abstract}

\section{Introduction}
\label{introduction}

Reading comprehension is a type of question answering
task where the answer is to be found in a passage about particular
entities and events. In
particular, the entities and events should not be mentioned in
structured databases of general knowledge. Reading comprehension
problems are intended to measure a system's ability to extract semantic
information about entities and relations directly from unstructured
text. 

Several large scale reading comprehension datasets have been
introduced recently, including 
the CNN \& Daily Mail
datasets~\citep{DeepMind}, the Children's Book Test (CBT)~\citep{CBT},
and the Who-did-What dataset~\citep{wdw}. The large sizes of these
datasets enable the application of deep learning.  These are all
cloze-style datasets where a question is constructed by deleting a
word or phrase from an article summary (in CNN/Daily Mail), from a sentence in a children's story (in CBT), or by deleting
a person from the first sentence of a different news article on the
same entities and events (in Who-did-What).

In this paper we present empirical evidence for the emergence of predication structure in a certain class of neural readers.
To understand predication structure, it is helpful to review the anonymization performed in the CNN/Daily Mail dataset.  In this dataset named entities are replaced by anonymous
entity identifiers such as ``entity37''. The passage might contain ``entity52 gave entity24 a rousing applause'' and the question might be ``$X$ received a rounding applause from entity52''.
The task is to fill in $X$ from a given multiple choice list of candidate entity identifiers. A fixed relatively small set of the same entity identifiers are used
over all the problems and the same problem is presented many times with different entity identifiers shuffled. This prevents a given entity identifier
from having any semantically meaningful vector embedding. The embeddings of the entity identifiers are presumably just pointers to semantics-free tokens.
We will write entity identifiers as logical constant symbols such as $c$ rather than strings such as ``entity37''.

``Aggregation'' readers, including Memory Networks \citep{firstmem,mem}, the Attentive Reader \citep{DeepMind}, 
and the Stanford Reader \citep{stanforder}, use bidirectional LSTMs or GRUs to construct a contextual embedding $h_t$ of each position $t$ in the passage
and also an embedding $h_q$ of the question $q$.  They then select an answer $c$ using a criterion similar to
\begin{equation}
  \label{eqn1}
  \argmax_c  \; \sum_t\;   <h_t,h_{q}>\;\;\; <h_t,e(c)>
\end{equation}
where $e(c)$ is the vector embedding of the constant symbol (entity identifier) $c$.  In practice the inner-product $<h_t,h_{q}>$ is normalized over $t$
using a softmax to yield attention weights $\alpha_t$ over $t$ and (\ref{eqn1}) becomes
\begin{equation}
  \label{eqn2}
  \argmax_c  \; <e(c),\;\sum_t\; \alpha_t h_t>.
\end{equation}
Here $\sum_t\;\alpha_t h_t$ can be viewed as a vector representation of the passage.

We argue that for aggregation readers, roughly defined by (\ref{eqn2}), the hidden state $h_t$ of the passage at position (or word) $t$ can be viewed as a vector concatenation
$h_t = [s(\Phi_t), s(c_t)]$ where $\Phi_t$ is a property (or statement or predicate) being stated of a particular constant symbol $c_t$.  Here $s(\Phi_t)$ and $s(c_t)$ are unknown emergent embeddings
of $\Phi_t$ and $c_t$ respectively.  
A logician might write this as $h_t = \Phi_t[c_t]$.
Furthermore, the question can be interpreted as having the form $\Psi[x]$ where the problem is to find a constant symbol $c$ such that the passage implies $\Psi[c]$.
Assuming $h_t = [s(\Phi_t),s(c_t)]$, $h_{q} = [s(\Psi),0]$, and $e(c) = [0,s(c)]$, we can rewrite (\ref{eqn1}) as
\begin{equation}
  \label{eqn3}
  \argmax_c   \; \sum_t \;    <s(\Phi_t),s(\Psi)>\;\;\;  <s(c_t),s(c)>.
\end{equation}
The first inner product in (\ref{eqn3}) is interpreted as measuring the extent to which $\Phi_t[x]$ implies $\Psi[x]$ for any $x$.
The second inner product is interpreted as restricting $t$ to positions talking about the constant symbol $c$.

Note that the posited decomposition of $h_t$ is not explicit in (\ref{eqn2}) but instead must emerge during training.  We present empirical evidence
that this structure does emerge.  The empirical evidence is somewhat tricky as the direct sum structure that divides $h_t$ into its two parts need not be axis
aligned and therefore need not literally correspond to vector concatenation.

We also consider a second class of neural readers that we call ``explicit reference'' readers.
Explicit reference readers avoid (\ref{eqn2}) and instead use
\begin{equation}
  \label{eqn4}
  \argmax_c \sum_{t \in R(c)} \alpha_t
\end{equation}
where $R(c)$ is the subset of the positions where the constant symbol (entity identifier) $c$ occurs.  Note that if we identify $\alpha_t$
with $<s(\Phi_t),s(\Psi)>$ and assume that $<s(c),s(c_t)>$ is either 0 or 1 depending on whether $c = c_t$, then (\ref{eqn3}) and (\ref{eqn4}) agree.
In explicit reference readers the hidden state $h_t$ need not carry a pointer to $c_t$ as the restriction on $t$ is independent of learned representations.
Explicit reference readers include the Attention Sum Reader \citep{AS},
the Gated Attention Reader \citep{GA}, and the Attention-over-Attention Reader \citep{AoA}.

So far we have only considered anonymized datasets that require the handling of semantics-free constant symbols.  However, even for non-anonymized
datasets such as Who-did-What, it is helpful to add features which indicate which positions in the passage are referring to which candidate answers.
This indicates, not surprisingly, that reference is important in question answering.  The fact that explicit reference features are needed
in aggregation readers on non-anonymized data indicates that reference is not being solved by the aggregation readers.  However, as reference seems to be important
for cloze-style question answering, these problems may ultimately provide training data from which reference resolution can be learned.

Sections~\ref{related_work} and~\ref{related_model} review various existing datasets and models respectively.
In the CNN dataset the vector embeddings of entity identifiers such as ``entity32'' are clearly interpretable as vector representations of semantics-free constant symbols.
However, to the best of our knowledge the emergent decomposition of the hidden state vectors into a concatenation of a property vector and an entity vector has not been previously described
or empirically investigated in the literature.
Section~\ref{analysis} presents
the logical structure interpretation of aggregation readers in more detail and
the empirical evidence supporting it.
Section~\ref{sec:reference} proposes new models that enforce the direct sum
structure of the hidden state vectors.  It is shown that these new models perform well on the Who-did-What
dataset provided that reference annotations are added as input features.
Section~\ref{sec:reference} also describes additional linguistic features that can be added to
the input embeddings and show that these improve the performance of existing models resulting in the best single-model performance
to date on the Who-did-What dataset.

\section{A Brief Survey of Datasets}
\label{related_work}

Before presenting various models for machine comprehension we give a general formulation of the machine comprehension task.
We take an instance of the task to be a four tuple $(q, p, a, \mathcal{A})$, where $q$ is a question given as a sequence of words containing a special token for a ``blank'' to be filled in,
$p$ is a document consisting of a sequence of words, $\mathcal{A}$ is a set of possible answers and $a \in \mathcal{A}$ is the ground truth answer.
All words are drawn from a vocabulary $\mathcal{V}$.  We assume
that all possible answers are words from the vocabulary, that is
$\mathcal{A} \subseteq \mathcal{V}$, and that the ground truth answer appears in the document, that is
$a \in p$. The problem can be described as that of selecting the answer $a \in \mathcal{A}$ that answers question $q$ based on information from $p$. 
We now briefly summarize important features of the related datasets in reading comprehension. 

\textbf{CNN \& Daily Mail}: \citet{DeepMind} constructed these datasets from a large number of news articles from the CNN
and Daily Mail news websites. The main article is used as the context, while the cloze style question is formed from one
short article summary sentence appearing in conjunction with the published article. To avoid the model using external world
knowledge when answering the question, the named entities in the entire dataset were replaced by anonymous entity IDs
which were then further shuffled for each example. This forces models to rely on the context document to answer each
question. In this anonymized corpus the entity identifiers are taken to be a part of the vocabulary and the answer set ${\cal A}$
consists of the entity identifiers occurring in the passage.

\textbf{Who-did-What (WDW)}: The Who-did-What dataset~\citep{wdw} contains 127,000 multiple choice cloze questions
constructed from the LDC English Gigaword newswire corpus~\citep{ldc}. In contrast with CNN and Daily Mail, WDW avoids using
article summaries for question formation. Instead, each problem is formed from two independent articles: one is given as
the passage to be read and a different article on the same entities and events is used to form the question. Further,
WDW avoids anonymization --- each choice is a person named entity. In this dataset the answer set ${\cal A}$
consists of the person named entities occurring in the passage. Finally, the problems have been filtered to
remove a fraction that are easily solved by simple baselines. It has two training sets. The larger training set
(``relaxed'') is created using less baseline filtering, while the smaller training set (``strict'') uses the same
filtering as the validation and test sets.

{\bf Other Related Datasets.} It is also worth mentioning several related datasets. The MCTest dataset~\citep{mctest} consists of children's stories and questions written by crowdsourced workers. The dataset only contains 660 documents
and is too small to train deep models. The bAbI dataset \citep{babi} is constructed automatically using synthetic text generation and
can be perfectly answered by hand-written algorithms~\citep{rvs}. The SQuAD dataset~\citep{squad} consists of  passage-question pairs 
where the passage is a Wikipedia article and the questions are written via crowdsourcing. The dataset contains over 100,000 problems, but the answer is often a word sequence which is difficult to handle with the reader models considered here. The Children's Book Test (CBT)~\citep{CBT} takes any sequence of 21 consecutive sentences from a children's book: the first 20 sentences are used as the passage, and the goal is to infer a missing word in the 21st sentence. The task complexity varies with the type of the omitted word (verb, preposition, named entity, or common noun). 
The LAMBADA dataset~\citep{lambada} is a word prediction dataset which requires
a broad discourse context, though the correct answer might not actually be contained in the context. Nevertheless, when the correct answer is in the
context, neural readers can be applied effectively~\citep{lambada_zewei}. 

\section{Aggregation Readers and Explicit Reference Readers}
\label{related_model}

As outlined in the introduction, here we classify readers into aggregation readers and explicit reference readers. Aggregation readers appeared first
in the literature and include Memory Networks~\citep{firstmem,mem}, the Attentive Reader~\citep{DeepMind}, and the Stanford Reader~\citep{stanforder}.
In this section we define aggregation readers more specifically by equations (\ref{eqn:r}) and (\ref{eqn:testloss}) below.  Explicit reference readers
include the Attention-Sum Reader~\citep{AS}, the Gated-Attention Reader~\citep{GA}, and the Attention-over-Attention Reader~\citep{AoA}. In this
section we define explicit reference readers more specifically by equation (\ref{eqn:testloss-AS}) below.  We first present the Stanford Reader as a paradigmatic
aggregation reader and the Attention-Sum Reader as a paradigmatic explicit reference reader.

\subsection{Aggregation Readers}

{\bf Stanford Reader.} The Stanford Reader \citep{stanforder} computes a bidirectional LSTM representation of both the passage and the question.
\begin{eqnarray}
  \label{eqn:h}
  h & = & \text{biLSTM}(e(p)) \\
  \label{eqn:q}
  h_{q} & = & [\text{fLSTM}(e(q))_{|q|},\text{bLSTM}(e(q))_1]
\end{eqnarray}
In equations (\ref{eqn:h}) and (\ref{eqn:q}) we have that $e(p)$ is the sequence of word embeddings $e(w_i)$ for $w_i \in p$ and similarly for $e(q)$.  The expression $\text{biLSTM}(s)$ denotes the sequence of hidden state vectors resulting from running
a bidirectional LSTM on the vector sequence $s$.  We write $\text{biLSTM}(s)_i$ for the $i$th vector in this sequence.  Similarly $\text{fLSTM}(s)$ and $\text{bLSTM}(s)$ denote the sequence of vectors resulting from running a forward LSTM
and a backward LSTM respectively and $[\cdot,\cdot]$ denotes vector concatenation.
The Stanford Reader, and various other readers, then compute a bilinear attention over the passage which is used to construct a single weighted vector representation of the passage. 
\begin{eqnarray}
  \label{eqn:attention}
  \alpha_{t}  =  \softmax_{t} \;h_{t}^\top W_\alpha \;h_{q} \quad
  \label{eqn:r}
  o  =  \sum_t \alpha_t h_t
\end{eqnarray}
Finally, they compute a probability distribution $P$ over the answers:
\begin{eqnarray}
  \label{eqn:trainloss}
  P(\cdot|d,q,{\cal A}) & = & \softmax_{a \in {\cal A}} \; e_o(a)^{\top}o \\
  \label{eqn:testloss}
  \hat{a} & = & \argmax_{a \in \mathcal{A}}\; e_o(a)^\top o
\end{eqnarray}
Here $e_o(a)$ is the ``output embedding'' of the answer $a$.  On the CNN dataset the Stanford Reader trains an output embedding for each of the roughly 550 entity identifiers used in the dataset.
For datasets in which the answer might be any word in ${\cal V}$, output embeddings must be trained for the entire vocabulary.

The reader is trained with log-loss $-\log P(a|p,q,{\cal A})$ where $a$ is the correct answer. At test time the reader
is scored on the percentage of problems where $\hat{a} = a$.

{\bf Memory Networks.} Memory Networks \citep{firstmem,mem} use (\ref{eqn:r}) and (\ref{eqn:testloss}) but have more elaborate methods of
constructing ``memory vectors'' $h_t$ not involving LSTMs.  Memory networks use (\ref{eqn:r}) and (\ref{eqn:testloss}) but replace (\ref{eqn:trainloss})
with
\begin{equation}
  \label{eqn:trainloss-wide}
  P(\cdot|p,q,{\cal A}) = P(\cdot|p,q) = \softmax_{w \in {\cal V}} e_o(w)^\top o.
\end{equation}
It should be noted that (\ref{eqn:trainloss-wide}) trains output vectors over the whole vocabulary rather than just those items occurring in the choice set ${\cal A}$.
This is empirically significant in non-anonymized datasets such as CBT and Who-did-What where choices at test time may never have occurred as choices in the training data.

{\bf Attentive Reader.} The Stanford Reader was derived from the Attentive Reader \citep{DeepMind}.
The Attentive Reader uses $\alpha_t = \softmax_t \text{MLP}([h_t,h_{q}])$
instead of (\ref{eqn:attention}).
Here $\text{MLP}(x)$ is the output of a multi layer perceptron given input $x$.
Also, the answer distribution in the Attentive Reader is defined over the full vocabulary rather than just the candidate answer set ${\cal A}$:
\begin{equation}
  \label{eqn:trainloss-wide2}
P(\cdot|p,q,{\cal A}) = \softmax_{w \in {\cal V}}\: e_o(w)^\top \text{MLP}([o,h_{q}])
\end{equation}
Equation (\ref{eqn:trainloss-wide2}) is similar to (\ref{eqn:trainloss-wide}) in that it leads to the training of output vectors for the full vocabulary
rather than just those items appearing in choice sets in the training data.  As in memory networks, this leads to improved performance on non-anonymized datasets.

\subsection{Explicit Reference Readers}

{\bf Attention-Sum Reader.} In the Attention-Sum Reader \citep{AS}, $h$ and $q$ are computed
with equations (\ref{eqn:h}) and (\ref{eqn:q}) as in the Stanford Reader
but using GRUs rather than LSTMs.
The attention $\alpha_t$ is computed similarly to (\ref{eqn:attention}) but using a simple inner product $\alpha_{t} = \softmax_{t} \;h_{t}^\top h_{q}$
rather than a trained bilinear form.  Most significantly, however, equations (\ref{eqn:trainloss}) and (\ref{eqn:testloss}) are replaced by
the following where $t \in R(a,p)$ indicates that a reference to candidate answer $a$ occurs at position $t$ in $p$.
\begin{eqnarray}
  \label{eqn:trainloss-AS}
  P(a|p,q,{\cal A}) & = & \sum_{t \in R(a,p)} \;\alpha_t \\
  \label{eqn:testloss-AS}
  \hat{a} & = & \argmax_a \sum_{t \in R(a,p)} \;\alpha_t 
\end{eqnarray}
Here we think of $R(a,p)$ as the set of references to $a$ in the passage $p$.  It is important to note that (\ref{eqn:trainloss-AS}) is an equality
and that $P(a|p,q,{\cal A})$ is not normalized to the members of $R(a,p)$.  When training with the log-loss objective this drives the attention $\alpha_t$
to be normalized --- to have support only on the positions $t$ with $t \in R(a,p)$ for some $a$.  See the heat maps in the supplementary material.

\textbf{Gated-Attention Reader.} The Gated-Attention Reader \citep{GA} involves a $K$-layer biGRU architecture defined by the following equations.
\begin{align*}
  h_{q}^{\ell} & = [\text{fGRU}(e(q))_{|q|},\text{bGRU}(e(q))_{1}] \;\;1 \leq \ell \leq K\\
  h^1 & = \text{biGRU}(e(p)) \\
  h^{\ell} & = \text{biGRU}(h^{\ell-1} \odot h_{q}^{\ell-1})\;\;2 \leq \ell \leq K
\end{align*}
Here the question embeddings $h_{q}^\ell$ for different values of $\ell$ are computed with different GRU model parameters.  Here $h \odot h_{q}$ abbreviates the sequence $h_1\odot h_{q}$, $h_2 \odot h_{q}$, $\ldots$ $h_{|p|}\odot h_{q}$.
Note that for $K=1$ we have only $h_{q}^1$ and $h^1$ as in the attention-sum reader.
An attention is then computed over the final layer $h^K$ with $\alpha_{t} = \softmax_{t} \;(h^K_{t})^\top \;h_{q}^{K}$ in the Attention-Sum Reader.
This reader uses (\ref{eqn:trainloss-AS}) and (\ref{eqn:testloss-AS}).

\textbf{Attention-over-Attention Reader.} The Attention-over-Attention Reader \citep{AoA} uses a more elaborate method to compute the attention $\alpha_t$.
We will use $t$ to range over positions in the passage and $j$ to range over positions in the question.  The model is then defined by the following equations.
$$\begin{array}{cc}
  h = \text{biGRU}(e(p)) & h_{q} = \text{biGRU}(e(q)) \\
  \\
  \alpha_{t,j} = \softmax_t h_t^\top h_{q,j} &\beta_{t,j} = \softmax_j h_t^\top h_{q,j} \\
  \\
\beta_j = \frac{1}{|p|} \sum_t \beta_{t,j} & \alpha_t = \sum_j \beta_j \alpha_{t,j}
\end{array}$$
Note that the final equation defining $\alpha_t$ can be interpreted as applying the attention $\beta_j$ to the attentions $\alpha_{t,j}$.
This reader uses (\ref{eqn:trainloss-AS}) and (\ref{eqn:testloss-AS}).

\section{Emergent Predication Structure}
\label{analysis}
      
As discussed in the introduction the entity identifiers such as ``entity37'' introduced in the CNN/Daily Mail datasets cannot be assigned any
semantics other than their identity.  We should think of them as pointers
or semantics-free constant symbols.  Despite this undermining of semantics, aggregation readers using (\ref{eqn:r}) and (\ref{eqn:testloss})
are able to perform well.  Here we posit that this is due to an emergent predication structure in the hidden vectors $h_t$.
Intuitively we want to think of the hidden state vector $h_t$ as a concatenation $[s(\Phi_t),s(a_t)]$ where $\Phi_t$ is a property being asserted
of entity $a_t$ at the positon $t$ in the passage.  Here $s(\Phi_t)$ and $s(a_t)$ are emergent embeddings of the property and entity respectively,
We also think of the vector representation $q$ of the question as having the form $[s(\Psi),0]$ and the vector embedding $e_o(a)$ as having the form
$[0,s(a)]$.

Unfortunately, the decomposition of $h_t$ into this predication structure need not be axis aligned.  Rather than posit an axis-aligned concatenation
we posit that the hidden vector space $H$ is a possibly non-aligned direct sum
\begin{equation}
  \label{eqn:predict3}
  H = S \oplus E
\end{equation}
where $S$ is a subspace of ``statement vectors'' and $E$ is an orthogonal subspace of ``entity pointers''.
Each hidden state vector $h \in H$ then has a unique
decomposition as $h = \Psi + e$ for $\Psi \in S$ and $e \in E$.  This is equivalent to saying that the hidden vector space $H$ is some rotation
of a concatenation of the vector spaces $S$ and $E$.  In this non-axis aligned model we also assume emergent embeddings $s(\Phi)$ and $s(a)$ with $s(\Phi) \in S$
and $s(a) \in E$.  We will also assume that the latent spaces are learned in such a way that explicit entity output embeddings satisfy $e_o(a) \in E$.

We now present empirical evidence for this decomposition structure.  This structure implies
$e_o(a)^\top h_t$ equals $e_o(a)^\top s(a_t)$.
This suggests the following for some fixed positive constant $c$.
\begin{equation}
  \label{eqn:predict1}
  e_o(a)^\top h_t = \left\{\begin{array}{ll} c & \mbox{if}\;t \in R(a,p) \\
      0 & \mbox{otherwise}
    \end{array}\right.
\end{equation}
We note that if $e_o(a)^\top s(a)$ was different for different constant $a$ then
answers would be biased toward constant symbols where this product was larger.  But we need to have that all constant symbols are equivalent.
We note that (\ref{eqn:predict1}) gives
\begin{align*}
\argmax_a \;e_o(a)^\top o  =  \argmax_a\;e_o(a)^\top \sum_t\; \alpha_t h_t \\
=  \argmax_a\;\sum_t\; \alpha_t\; e_o(a)^\top h_t = \argmax_a \sum_{t \in R(a,p)} \alpha_t
\end{align*}
and hence (\ref{eqn:testloss}) and (\ref{eqn:testloss-AS}) agree --- the aggregation readers and the explicit reference readers are using essentially the same
answer selection criterion.

   \begin{table*}
   \begin{center}
     \begin{tabular}{@{}lcccccccc@{}}
       & \multicolumn{3}{c}{ CNN Dev} & \multicolumn{3}{c}{CNN Test} \\
       & samples&mean&variance&samples&mean&variance\\ \\
       $e_o(a)^\top h_t, \quad t \in R(a, p)$     & 222,001&10.66 &2.26 & 164,746&10.70 &2.45 \\
       $e_o(a)^\top h_t, \quad t \notin R(a, p)$ & 93,072,682&-0.57 &1.59 & 68,451,660&  -0.58 & 1.65 \\
       $e_o(a)^\top h_{t\pm 1}, \quad t \in R(a, p)$     & 443,878 &2.32 &1.79 & 329,366& 2.25 &1.84 \\
       \\
       $\mathrm{Cosine}(h_{q}, h_t),  \quad \exists a\;t \in R(a, p)$ & 222,001& 0.22 & 0.11 & 164,746& 0.22&0.12 \\
       $\mathrm{Cosine}(h_{q}, e_o(a)),\quad \forall a $ & 103,909& -0.03 & 0.04 & 78,411& -0.03& 0.04\\
       
              
     \end{tabular}
   \end{center}
   \caption{Statistics to support (\ref{eqn:predict1}) and (\ref{eqn:predict2}). These statistics are computed for the Stanford Reader.}
   \label{table: statistics}
   \end{table*}
   
\begin{figure}[t]
\centering
\includegraphics[width=.75\linewidth]{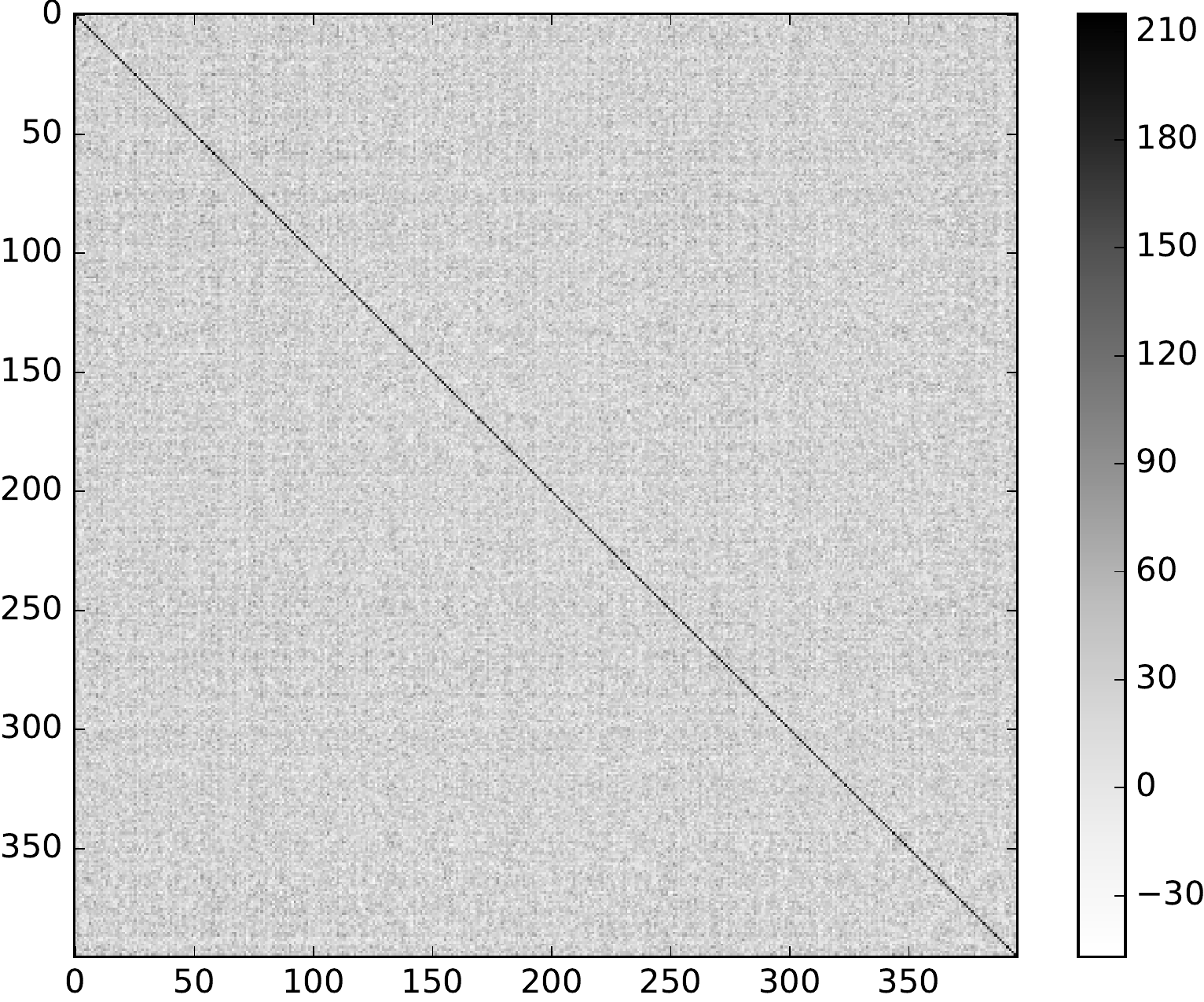}
\caption{Plot of $e_o(a_i)^\top e_o(a_j)$ from Stanford Reader trained on CNN dataset, where rows range over $i$ values and columns range over $j$ values. Off-diagonal values have mean 25.6 and variance 17.2 while diagonal values have mean 169 and variance 17.3.}
\label{fig:diag}
\end{figure}

Empirical evidence for (\ref{eqn:predict1}) is given in the first three rows
of Table~\ref{table: statistics}. The first row empirically measures the constant $c$ in (\ref{eqn:predict1}) by measuring $e_0(a)^\top h_t$
for those cases where $t \in R(a,p)$.  The second row measures ``0'' in (\ref{eqn:predict1}) by measuring $e_o(a)^\top h_t$ in those cases
where $t \not \in R(a,p)$.  The third row shows that this inner product falls off significantly just one word before or after the position of the
answer word. 
Additional evidence for (\ref{eqn:predict1}) is given in Figure~\ref{fig:diag} showing that the output vectors $e_o(a)$ for
different entity identifiers $a$ are nearly orthogonal. Orthogonality of the output vectors is required by (\ref{eqn:predict1}) provided that each output vector $e_o(a)$
is in the span of the hidden state vectors $h_{t,p}$ for which $t \in R(a,p)$.  Intuitively, the mean of all vectors $h_{t,p}$ with $t \in R(a,p)$
should be approximately equal to $e_o(a)$. Empirically this will only be approximately true.

Equation~(\ref{eqn:predict1}) would suggest that the vector embedding of the constant symbols should have dimension at least as large as the number of distinct constants. 
However, in practice it is sufficient that $e_0(a)^\top s(a')$ is small for $a \not = a'$.  This allows the vector embeddings of the constants to have dimension much smaller
than the number of constants.  We have experimented with two-sparse constant symbol embeddings where the number of embedding vectors in dimension $d$ is $2d(d-1)$ ($d$ choose 2
times the four ways of setting the signs of the non-zero coordinates).  Although we do not report results here, these designed and untrained constant embeddings worked reasonably well.

As further support for (\ref{eqn:predict1}) we give heat maps for $e_{o}(a)^\top h_{t}$ 
for different identifiers $a$ and heat maps for $\alpha_{t}$ for different readers in the supplementary material. 

As another testable predication we note that the posited decomposition of the hidden state vectors implies
\begin{equation}
  \label{eqn:predict2}
  h_{q}^\top(h_i + e_o(a)) = h_{q}^\top h_i.
\end{equation}
This equation is equivalent to $h_{q}^\top e_o(a) = 0$.  Experimentally, however, we cannot expect $h_{q}^\top e_o(a)$ to be exactly zero
and (\ref{eqn:predict2}) seems to provides a more experimentally meaningful test.  
Empirical evidence for (\ref{eqn:predict2}) is given in the fourth and fifth rows of Table~\ref{table: statistics}.
The fourth row measures the cosine of the angle between the question vector $h_{q}$ and the hidden state $h_t$ averaged over passage positions $t$
at which some entity identifier occurs.
The fifth row measures the cosine of the angle between $h_{q}$ and $e_o(a)$ averaged over the entity identifiers $a$.

A question asks for a value of $x$ such that a statement $\Psi[x]$ is implied by the passage.
For a question $\Psi$ we might even suggest the following vectorial interpretation of entailment.
$$\Phi[x]\;\mbox{implies} \;\Psi[x] \;\;\; \mbox{iff}\;\;\;  \Phi^\top \Psi \geq ||\Psi||_1.$$
This interpretation is exactly correct if some of the dimensions of the vector space correspond to predicates,
$\Psi$ is a 0-1 vector representing a conjunction predicates, and $\Phi$ is also 0-1 on these dimensions indicating
whether a predicate is implied by the context.  Of course in practice one expects the dimension to be smaller than the number
of possible predicates.

\section{Pointer Annotation Readers}
\label{sec:reference}

It is of course important to note that anonymization provides reference information\----anonymization assumes that one can determine
coreference so as to replace coreferent phrases with the same entity identifier.  Anonymization allows the reference set $R(a,p)$
to be directly read off of the passage.  Still, an aggregation reader must learn to recover this explicit reference structure.

Aggregation readers can have difficulty when anonymization is not done.
The Stanford Reader achieves just better than $45\%$ on the Who-did-What dataset while the Attention-Sum
Reader can get near $60\%$ (see Table \ref{table:resultswdw}). But if we anonymize the Who-did-What dataset and then re-train the Stanford Reader, the accuracy jumps to near $65\%$.
Anonymization greatly reduces the number of output word embeddings $e_o(a)$ to be learned. We need to learn
only output embeddings for the relatively small number of entity identifiers needed for the question.  Anonymization suppresses the semantics
of the reference phrases and leaves only a semantics-free entity identifier.  This suppression of semantics may facilitate the separation of the hidden
state vector space $H$ into a direct sum $S \oplus E$ with $s(\Phi) \in S$ and  $e_o(a),s(a) \in E$.

A third, and perhaps more important effect of anonymization is to provide reference information. Anonymization explicitly marks positions of candidate answers and establishes coreference.
A natural question is whether this information can be provided without anonymization by simply adding additional coreference features to the input.  Here we evaluate two architectures
inspired by this question.  This evaluation is done on the Who-did-What dataset which is not anonymized. In each architecture we add features to the input to mark the occurrences of candidate answers.
These models are simpler than the Stanford Reader but perform comparably.  This comparable performance in Table \ref{table:resultswdw} further supports our analysis of logical structure in aggregation readers.\\
         
\noindent \textbf{One-Hot Pointer Reader}: The Stanford Reader uses input embeddings of words and output embeddings of entity identifiers.
In the Who-did-What dataset each problem has at most five choices in the multiple choice answer list.  This means that we need only five entity identifiers
and we can use a five dimensional one-hot vector representation for answer identifiers. 

If an answer choice exists at position $t$ in the passage let $i_t$ be the index of that choice on the answer choice list.  If no answer choice occurs at position $t$ we let $i_t$ be zero.
We define $e'(i)$ to be the zero vector if $i=0$ and otherwise to be the one-hot vector for $i$ (i.e., the five-dimensional vector with zeroes at all positions except with a one at position $i$). 
We define ``pointer annotation'' to be the result of concatenating $e'(i_t)$ as additional features to the word embedding $e(w_t)$ for token $w_t$ in the passage:
\begin{equation}
  \label{eqn:pointer}
  \bar{e}(w_t) = [e(w_t),e'(i_t)]
  \end{equation} 
We feed the new $\bar{e}(w_t)$ to the readers for each token $w_t$. We define a ``one-hot pointer reader'' by designating the last five dimensions of the hidden state as indicators of the answer and take the probability of choice $i$ to be defined as
\begin{equation}
  \label{eqn:answer}
  p(i|d,q) = \softmax_{i \in \mathcal{A}}\; o_{i}
  \end{equation}
  where $o$ is computed by (\ref{eqn:r}) and $o_i$ is the $i$th-to-last dimension of vector $o$. Table \ref{table:resultswdw} shows results using this reader, showing performance comparable to the Stanford Reader with anonymization. \\
  
\noindent \textbf{General Pointer Reader}: In the CNN dataset there are roughly 550 entity identifiers and a one-hot representation may not be desirable because it would enlarge the embedding space too much.
  Instead we can let $e'(i)$ be a fixed set of ``pointer vectors''\----vectors distributed widely on the unit sphere so that for $i \not = j$ we have that $e'(i)^\top e'(j)$ is small.  We again use (\ref{eqn:pointer}) but replace (\ref{eqn:answer}) with
\begin{equation}
p(i|d,q) = \softmax_i\; [0,e'(i)]^{\top}o
\end{equation} 
\noindent where ``$0$'' stands for a sufficient number of zeroes in order to make the dimensions match. We refer to this as a ``general pointer reader''. In this reader, the pointer embeddings $e'(i)$ are held fixed and not trained. 
Even though not shown here, in preliminary experiments, this reader yield similar performance to the one hot pointer reader while permitting smaller embedding dimensionality.

\noindent \textbf{Linguistic Features}:  Each model can be modified to include additional input features for each input token in the question and passage.
More specifically we can add the following features to the word embeddings: whether the current token occurs in the question; the frequency of the current token in the passage; the position of the token's first occurrence in the passage as a percentage of the passage length; and whether the text surrounding the token matches the text surrounding the placeholder in the question. More details of the experimental setup are provided in the appendix.

Table \ref{table:resultswdw} shows results when adding these features to the Gated-Attention Reader, Stanford Reader, and One-Hot Pointer Reader, showing large improvements to all readers and leading to the best single-model performance reported to-date on the Who-did-What dataset. 

\begin{table*}
\begin{center}
\begin{tabular}{@{}|l|c|c|@{}}
  \hline
Who did What  & \text{Validation} & \text{Test} \\
  \hline
Attention Sum Reader~\citep{wdw}        &59.8 &58.8 \\ 
Gated Attention Reader~\citep{wdw}     &60.3 &59.6  \\
NSE~\citep{umass}     & 66.5 & 66.2 \\ 
Gated Attention + Linguistic Features$^{+}$ & 72.2 & \bf{72.8} \\ \hline
Stanford Reader & 46.1 & 45.8 \\
Attentive Reader with Anonymization        &55.7 & 55.5 \\ 
Stanford Reader with Anonymization     & 64.8 & 64.5 \\ 
One-Hot Pointer Reader & 65.1 & 64.4 \\
One-Hot Pointer Reader + Linguistic Features$^{+}$ & 69.3 & 68.7 \\
Stanford with Anonymization + Linguistic Features$^{+}$    & 69.7 &  \textcolor{blue}{\textbf{69.2}} \\ \hline
  Human Performance        &- &84 \\ \hline 
\end{tabular}
\end{center}
\caption{Accuracy on Who-did-What dataset. Each result is based on a single model. Results for neural readers other than NSE are based on replications of those systems.  All models were trained on the relaxed training set which uniformly yields better performance than the restricted training set. The first group of models are explicit reference models and the second group are aggregation models. $+$ indicates anonymization with better reference identifier. \label{table:resultswdw}}
\end{table*}

\section{Discussion}
\label{conclusion}

Explicit reference architectures rely on reference resolution\----a specification of which phrases in the
given passage refer to candidate answers. Our experiments indicate that all existing readers
benefit greatly from this externally provided information.  Aggregation readers seem to demonstrate
a stronger learning ability in that they essentially learn to mimic explicit reference readers
by identifying reference annotation and using it appropriately.  This is done most clearly in the
pointer reader architectures.  Furthermore, we have argued for, and given experimental evidence for,
an interpretation of aggregation readers as learning emergent predication structure\----a factoring of neural
representations into a direct sum of a statement (predicate) representation and an entity (argument) representation.

At a very high level our analysis and experiments support a central role for reference resolution
in reading comprehension.  Automating reference resolution in neural models, and demonstrating its value
on appropriate datasets, would seem to be an important area for future research.

There is great interest in learning representations for natural language understanding.  The current state of the art in reading
comprehension is such that systems still benefit from externally provided linguistic features including externally
annotated reference resolution. It would be interesting to develop fully automated neural readers
that perform as well as readers using externally provided annotations. 


\subsection*{Acknowledgments}
We thank NVIDIA Corporation for donating GPUs used in this research.

\bibliography{acl2017}
\bibliographystyle{acl_natbib}

\newpage
\appendix

\section{Supplemental Material}
\label{sec:supplemental}

\subsection{Experiment Details}

   We implemented the neural readers using Theano~\citep{theano} and Blocks~\citep{block} and train them on a single NVIDIA Tesla K40 GPU.  Negative log-likelihood is employed as training criterion. We used stochastic gradient descent (SGD) with the Adam update rule~\citep{Adam} and set the learning rate to 0.0005. 
   
   For the Stanford Reader and One-Hot Pointer Reader, we use the Stanford Reader's default settings. 
For the Gated-Attention reader, the lookup table was initialized using pre-trained GloVe~\citep{glove} vectors.\footnote{\url{http://nlp.stanford.edu/data/glove.6B.zip}} 
Input to hidden state weights were initialized by random orthogonal matrices~\citep{orh} and biases were initialized to zero. 
Hidden to hidden state weights were initialized by identity matrices to force the model to remember longer information. 
To compute the attention weight, we use $\alpha_{t} = h_{t}^\top W_{\alpha}h_{q}$ and initialize $W_{\alpha} $ with random uniform distribution. We also use gradient clipping~\citep{gc} with a threshold of 10 and mini-batches of size 32. 

During training we randomly shuffle all examples within each epoch. To speed up training, we always pre-fetch 10 batches worth of examples and sort them according to document length as done by \citet{AS}. When using anonymization, we randomly reshuffle the entity identifier to match the procedure proposed by \citet{DeepMind}. 
	
During training we evaluate the accuracy after each epoch and stop training when the accuracy on the validation set starts decreasing. We tried limiting the vocabulary to the most frequent tokens but did not observe any performance improvement compared with using all distinct tokens as the vocabulary. Since part of our experiments need to check word embedding assignment issues, we finally use all the distinct tokens as vocabulary. To find the optimal embedding and hidden state dimension, we tried several groups of different combinations, and the optimal values were 200 and 384, respectively.  

When anonymizing the Who-did-What dataset, we can either use simple string matching to replace answers in the question and passage with entity identifiers, or we can use the Stanford named entity recognizer (NER)\footnote{\url{http://nlp.stanford.edu/software/CRF-NER.shtml}} to detect named entities and replace the answer named entities in the question and passage with entity identifiers. We found the latter to bring 2\% improvement compared with simple string matching.

\subsection{Heat Maps for Stanford Reader for Different Answer Candidates}

\begin{figure}[h!]
   \centering      
   \includegraphics[clip, trim=1.5cm 3cm 0.5cm 3cm, width=1.1\linewidth]{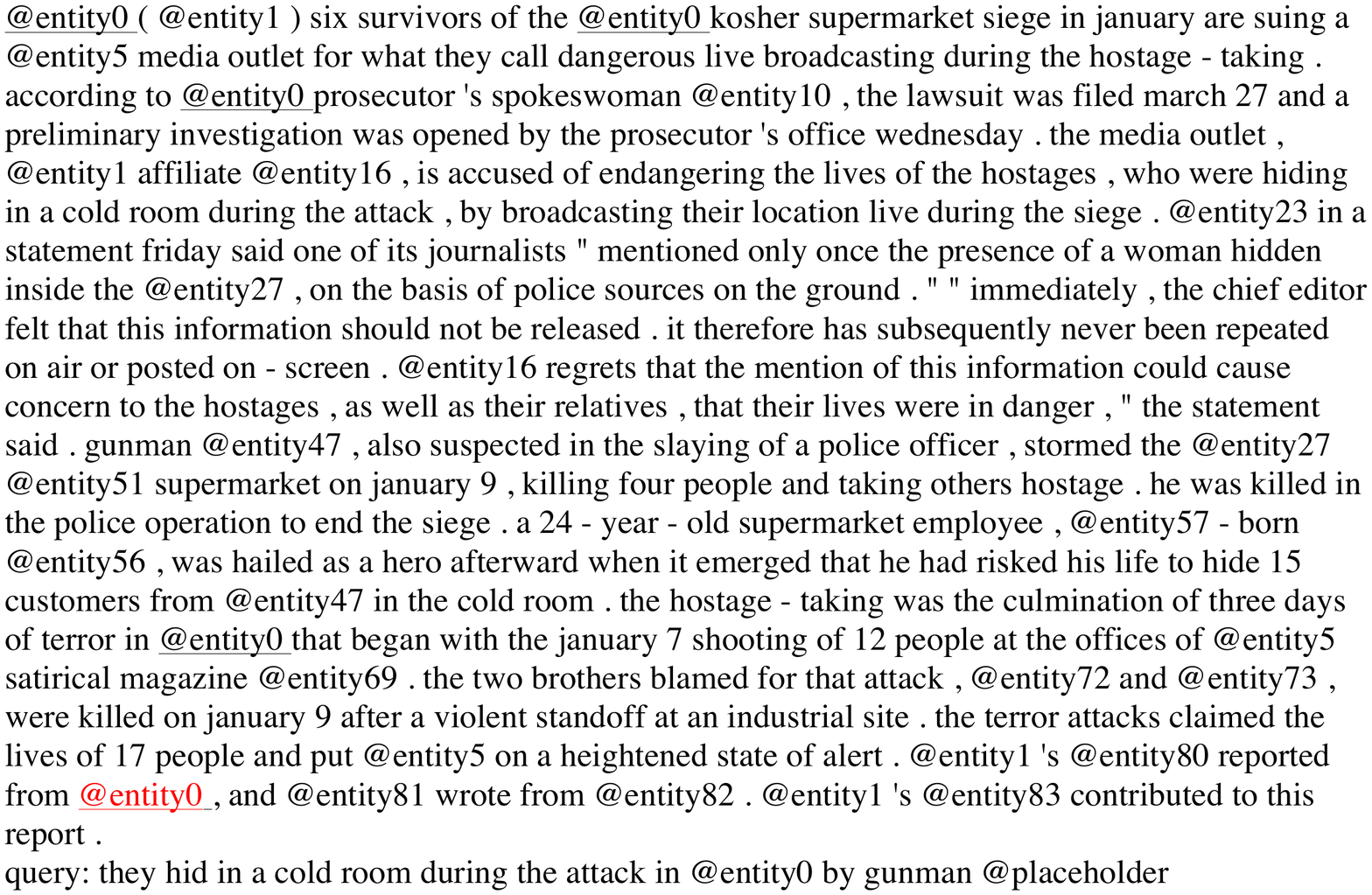}
 \caption{Heat map when $a$ = entity0.}
 \label{fig:SDReader:0}
\end{figure}

\begin{figure}[h!]
   \centering      
   \includegraphics[clip, trim=1.5cm 3cm 0.5cm 3cm, width=1.1\linewidth]{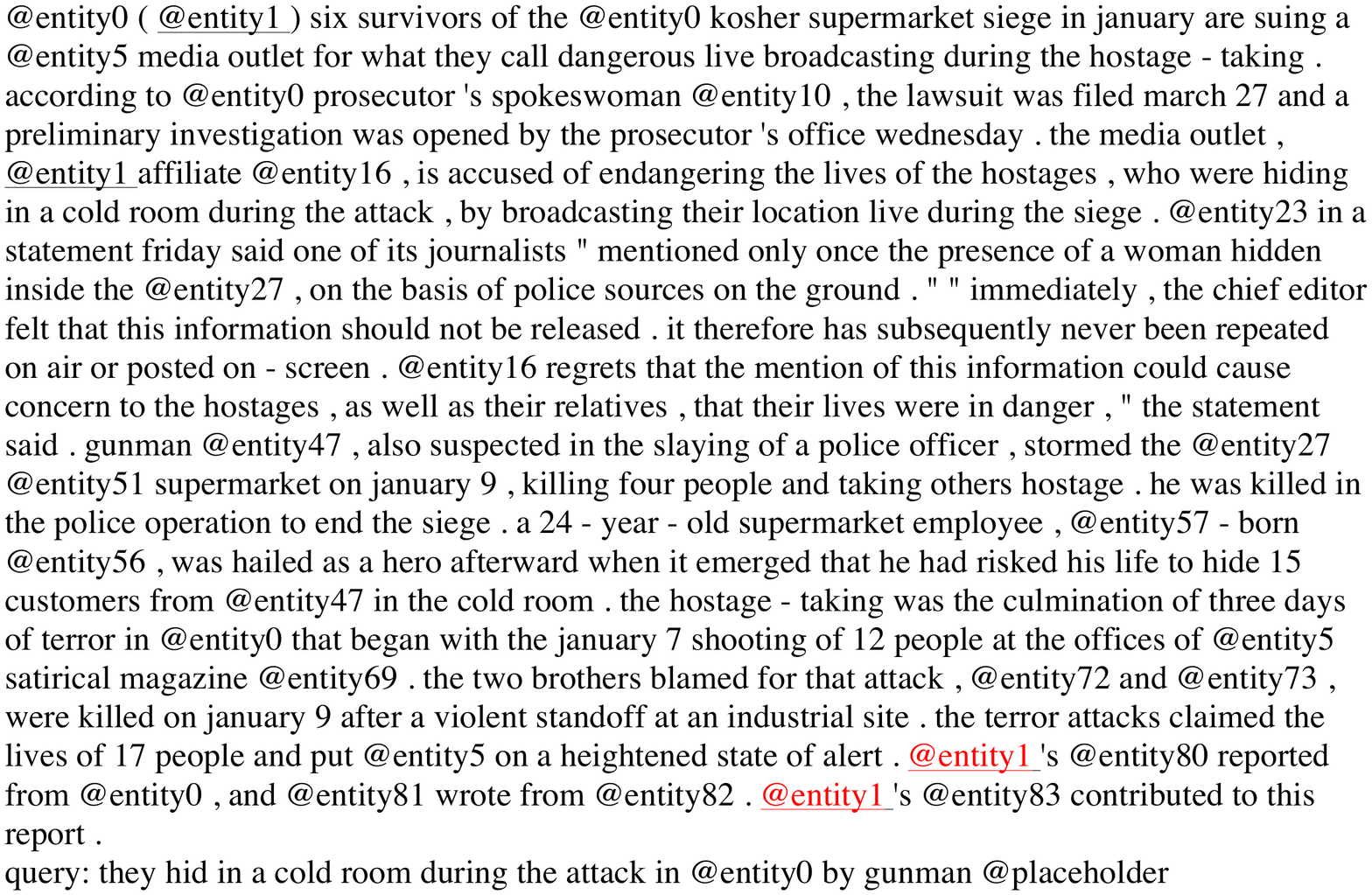}
 \caption{Heat map when $a$ = entity1.}
 \label{fig:SDReader:1}
\end{figure}

\begin{figure}[h!]
   \centering      
   \includegraphics[clip, trim=1.5cm 3cm 0.5cm 3cm, width=1.1\linewidth]{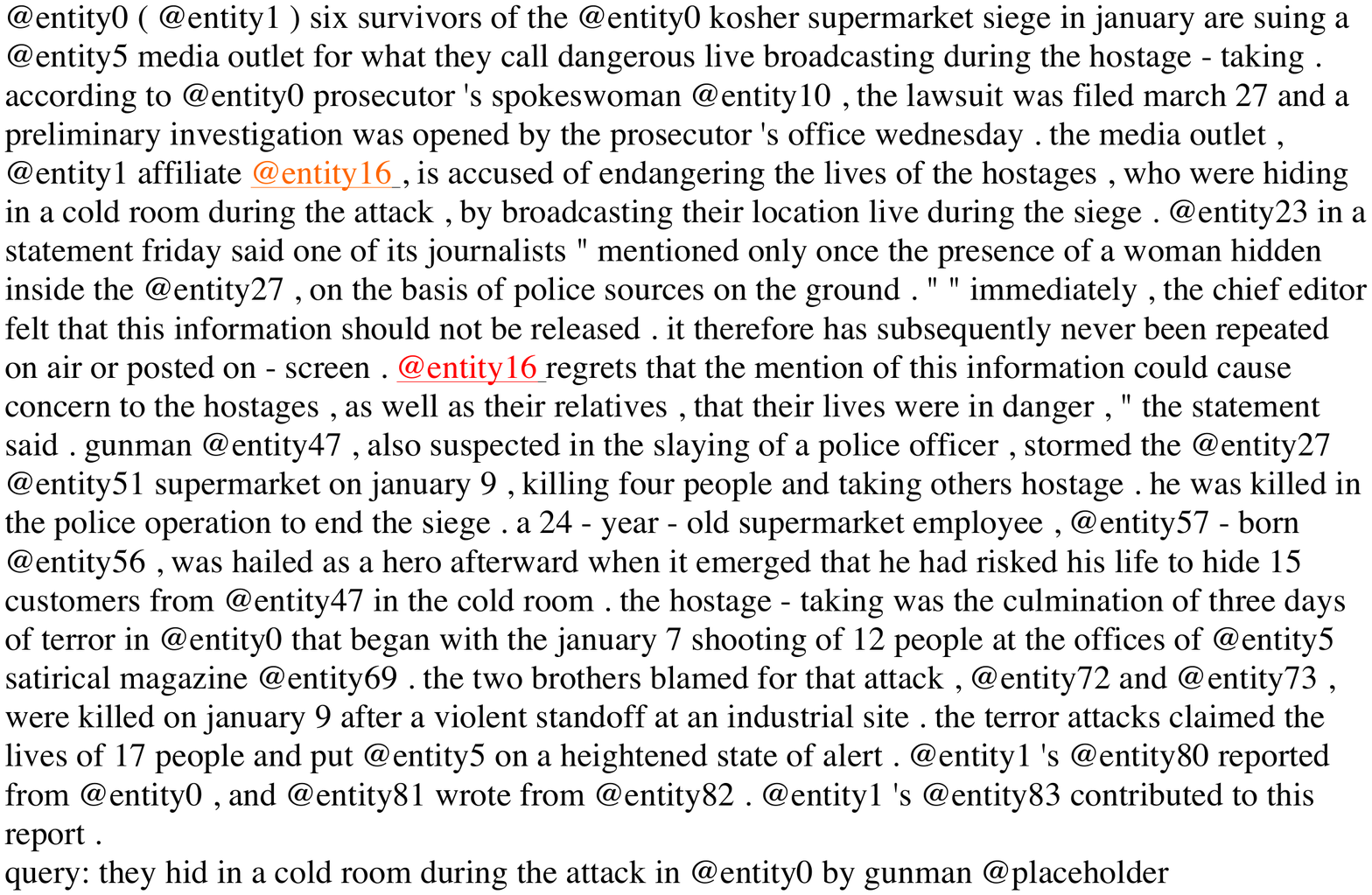}
 \caption{Heat map when $a$ = entity16.}
 \label{fig:SDReader:16}
\end{figure}

\begin{figure}[h!]
   \centering      
   \includegraphics[clip, trim=1.5cm 3cm 0.5cm 3cm, width=1.1\linewidth]{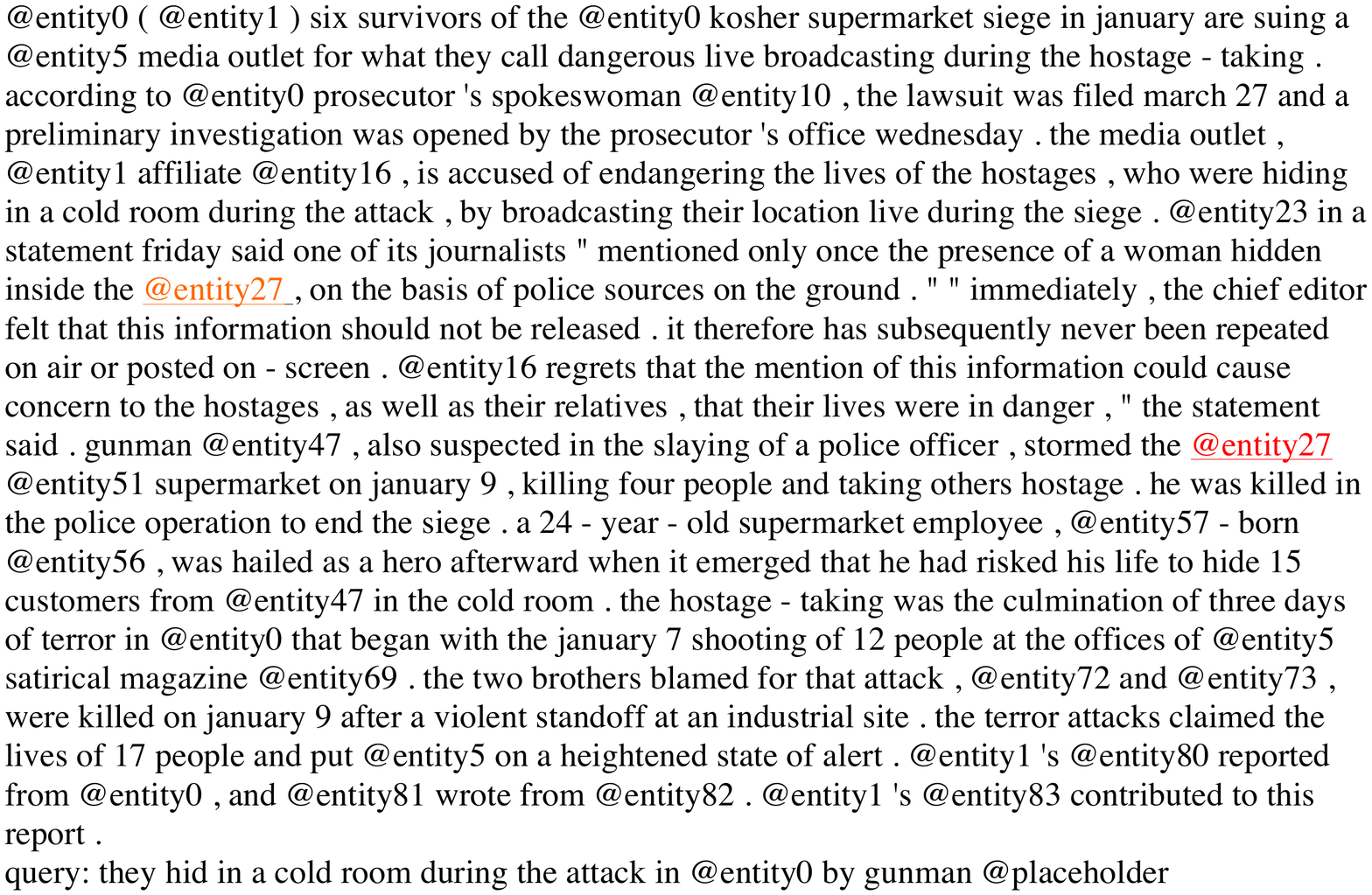}
 \caption{Heat map when $a$ = entity27.}
 \label{fig:SDReader:27}
\end{figure}

\begin{figure}[h!]
   \centering      
   \includegraphics[clip, trim=1.5cm 3cm 0.5cm 3cm, width=1.1\linewidth]{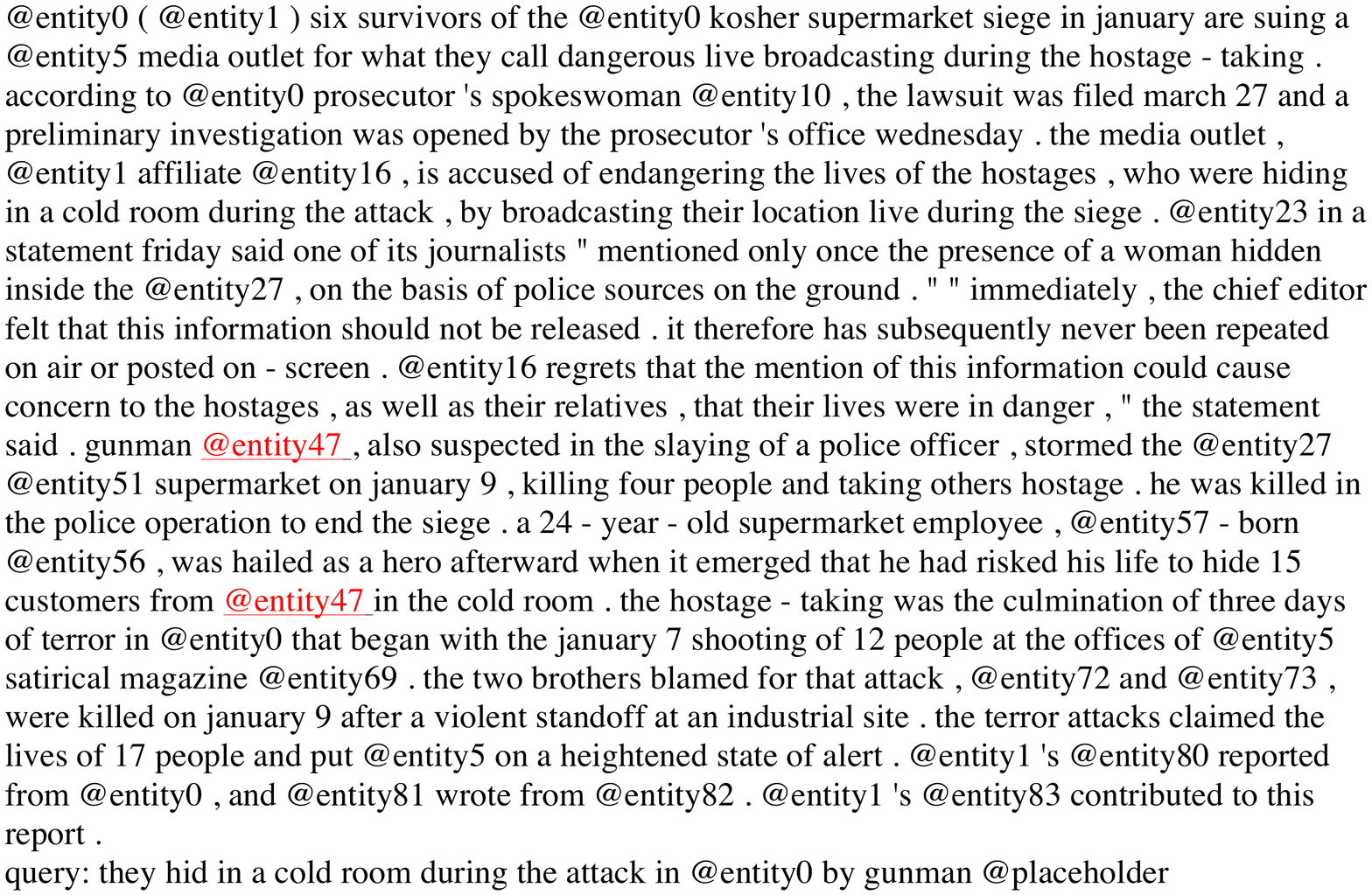}
 \caption{Heat map when $a$ = entity47.}
 \label{fig:SDReader:47}
\end{figure}

  We randomly choose one article from the CNN dataset and show $\softmax (e_{o}(a)^\top h_{t})$ for $t \in [0, |p|]$ for each answer candidate $a$ in Figures~\ref{fig:SDReader:0}-\ref{fig:SDReader:47}.
Red color indicates larger probability and orange indicates smaller probability and the remaining indicates very low probability that can be ignored. From these figures, we can see that our assumption that $e_{o}(a)$ is used to pick up its occurrence is reasonable.

\subsection{Heat Maps for Other Readers}

\begin{figure}[ht]
   \centering      
   \includegraphics[clip, trim=1.5cm 5cm 0.5cm 3.5cm, width=1.1\linewidth]{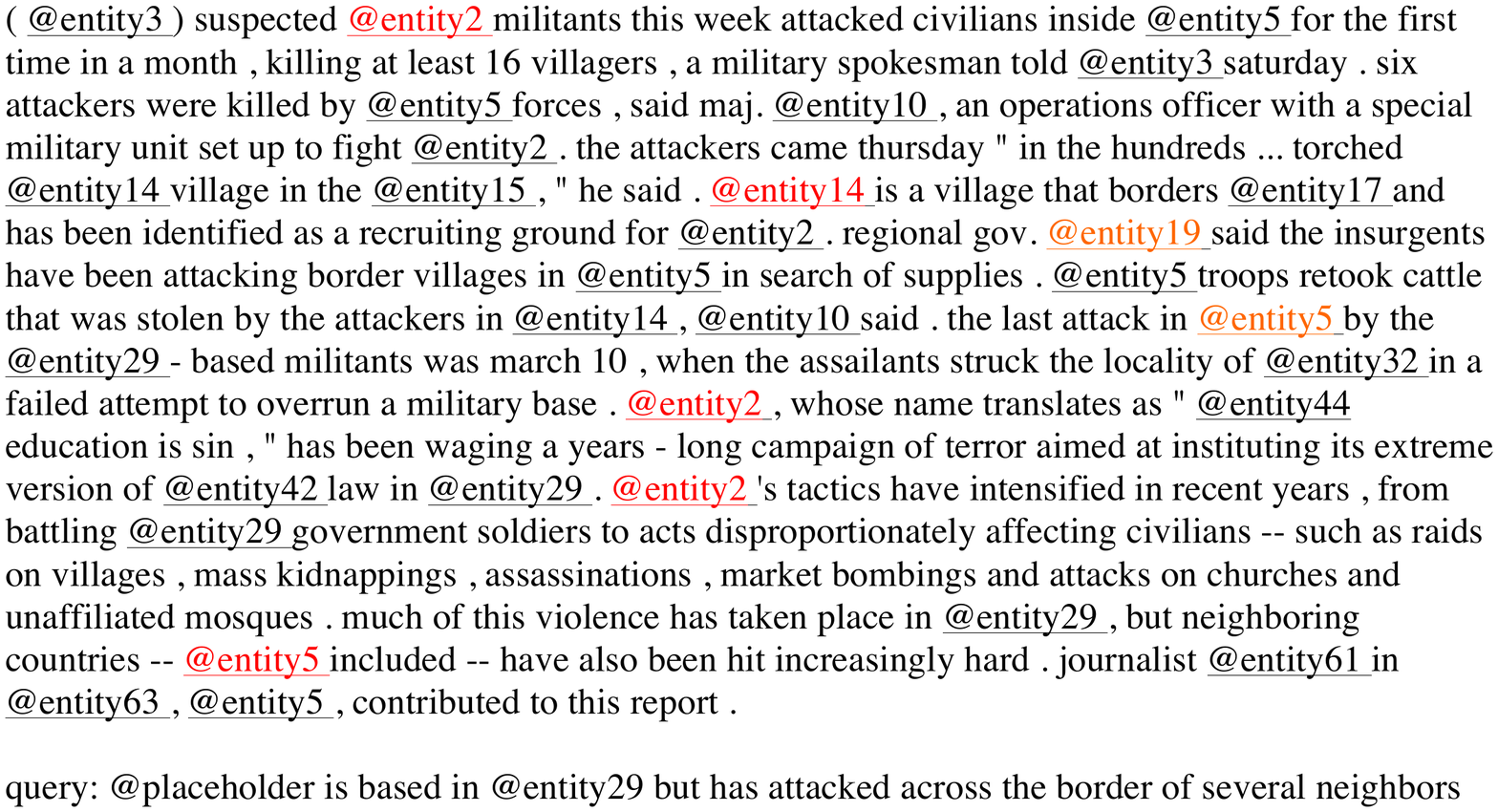}
 \caption{Heat map $\alpha_{t}$ for Stanford Reader.}
 \label{fig:SDReader}
\end{figure}

\begin{figure}[ht]
   \centering      
   \includegraphics[clip, trim=1.5cm 4cm 0.5cm 4cm, width=1.1\linewidth]{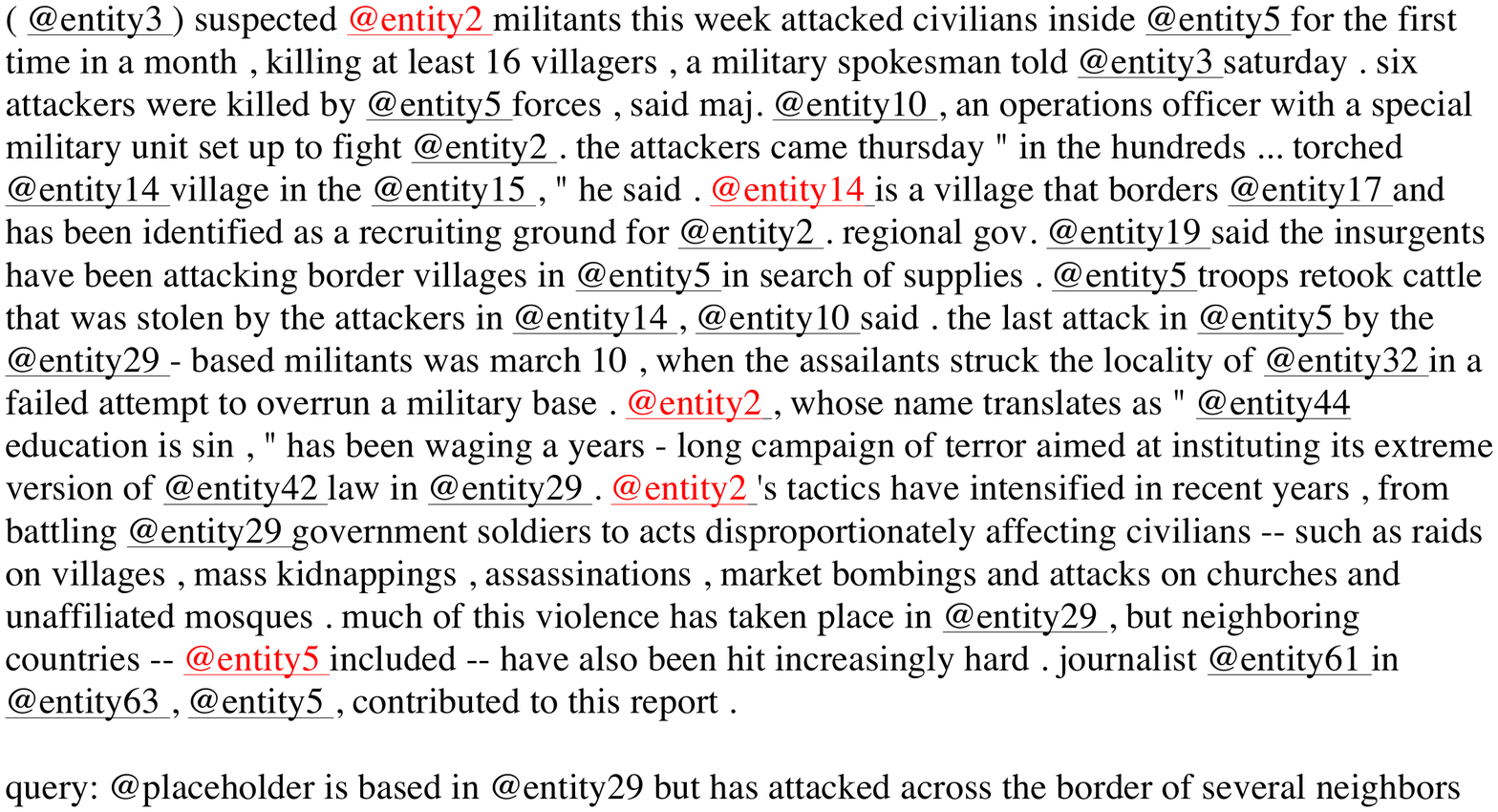} 
 \caption{Heat map $\alpha_{t}$ for Gated Attention Reader.}
 \label{fig:GAReader}
\end{figure}

\begin{figure}[ht]
   \centering      
   \includegraphics[clip, trim=1.5cm 5cm 0.5cm 3.5cm, width=1.1\linewidth]{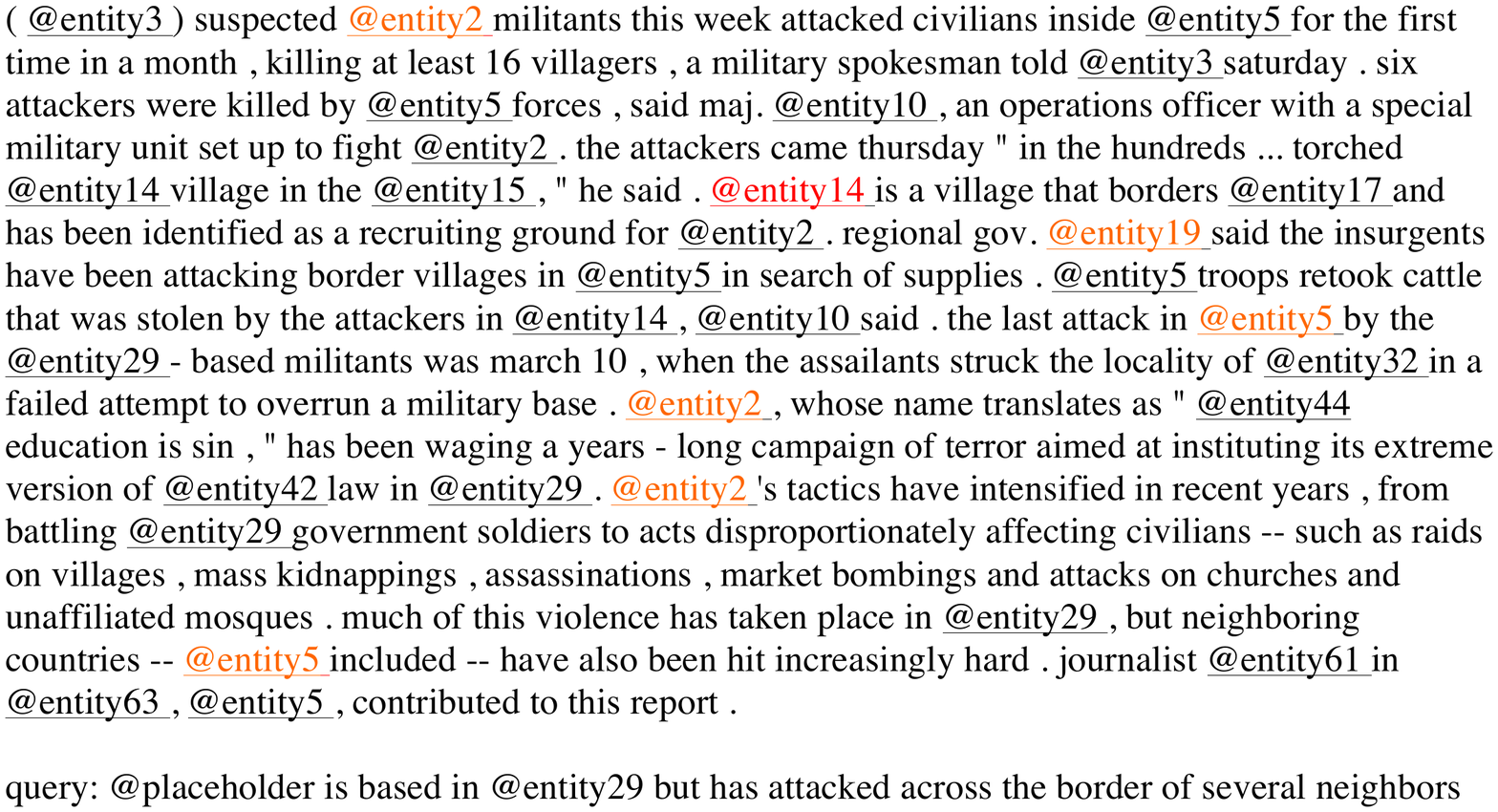} 
 \caption{Heat map $\alpha_{t}$ for Attention Sum Reader.}
 \label{fig:ASReader}
\end{figure}

We randomly choose one article from the CNN dataset and show the attention map $\alpha_{t} = \softmax (h_{q}^\top W_{a}h_{t})$ for different readers (in Attention Sum and Gated Attention Reader, $W_{\alpha}$ is identity matrix). 
In Figures~\ref{fig:SDReader}-\ref{fig:ASReader}, 
we can see that all readers put essentially all weight on the entity identifiers.

\end{document}